\documentclass[sigconf]{acmart}
\usepackage{listings}
\usepackage{xcolor}
\usepackage{amsfonts}
\usepackage{amsmath}
\usepackage{graphicx}
\usepackage[bb=dsserif]{mathalpha}
\usepackage{bm}
\usepackage{float}

\usepackage{xcolor}
\usepackage{booktabs}    
\usepackage{mathtools}
\usepackage{multirow}
\usepackage{caption}
\usepackage{subcaption}
\usepackage{tabularx}
\usepackage{amsmath,bm}
\usepackage{algorithm}
\usepackage{algpseudocode}
\usepackage{wrapfig}
\usepackage{lipsum}
\usepackage{tikz}
\def\checkmark{\tikz\fill[scale=0.4](0,.35) -- (.25,0) -- (1,.7) -- (.25,.15) -- cycle;} 

\setcopyright{rightsretained}

\usepackage{pifont}
\definecolor{codegreen}{rgb}{0,0.6,0}
\definecolor{codegray}{rgb}{0.5,0.5,0.5}
\definecolor{codepurple}{rgb}{0.58,0,0.82}
\definecolor{backcolour}{rgb}{0.95,0.95,0.92}

\usepackage{color, colortbl}
\usepackage[first=0,last=9]{lcg}

\definecolor{LightYellow}{rgb}{0.99, 0.99, 0.59}
\definecolor{LightRed}{rgb}{0.97, 0.51, 0.47}
\definecolor{LightBlue}{rgb}{0.99, 0.59, 0.99}
\definecolor{LightGreen}{rgb}{0.59, 0.99, 0.99}

\definecolor{codegreen}{rgb}{0,0.6,0}
\definecolor{codegray}{rgb}{0.5,0.5,0.5}

\definecolor{backcolour}{RGB}{245,248,250}
\definecolor{emph}{RGB}{166,88,53}
\definecolor{nightblue}{RGB}{9,49,105}
\definecolor{keywords}{RGB}{207,33,46}
\definecolor{lightpurple}{RGB}{130,81,223}

\lstdefinestyle{mystyle}{
    backgroundcolor=\color{backcolour},   
    commentstyle=\color{codegreen},
    keywordstyle=\color{keywords},
    stringstyle=\color{nightblue},
    basicstyle=\ttfamily\footnotesize,
    breakatwhitespace=false,         
    breaklines=true,                 
    captionpos=b,                    
    keepspaces=true,                 
    showspaces=false,                
    showstringspaces=false,
    showtabs=false,                  
    tabsize=2,
    frame=shadowbox,
    emph={AutoTokenizer,AutoModelForSequenceClassification,Explainer},
    emphstyle={\color{emph}},
    emph={[2]from_pretrained,compute_table},
    emphstyle={[2]\color{lightpurple}},
    linewidth=8.0cm
}

\usepackage{amsmath}
\usepackage{empheq}
\usepackage[most]{tcolorbox}
\usepackage{enumitem}

\newtcbox{\mymath}[1][]{%
    nobeforeafter, math upper, tcbox raise base,
    enhanced, colframe=blue!30!black,
    colback=blue!10, boxrule=1pt,
    #1}
    
\newtcbox{\mymathbox}[1][]{%
    nobeforeafter, math upper, tcbox raise base,
    enhanced, colframe=red!30!black,
    colback=red!10, boxrule=1pt,
    #1}

\newtcbox{\mymathgreen}[1][]{%
    nobeforeafter, math upper, tcbox raise base,
    enhanced, colframe=green!30!black,
    colback=green!10, boxrule=1pt,
    #1}

\usepackage[capitalize,noabbrev]{cleveref}

\copyrightyear{2024} 
\acmYear{2024} 
\setcopyright{rightsretained} 

\begin{document}

\title{Large Language Models As Evolution Strategies}

\author{Robert Tjarko Lange}
\affiliation{%
    \institution{TU Berlin, Google DeepMind}
    \country{Germany, Japan}
 }
\email{robert.t.lange@tu-berlin.de}

\author{Yingtao Tian}
\affiliation{%
    \institution{Google DeepMind}
    \country{Japan}
 }
\email{alantian@google.com}

\author{Yujin Tang}
\affiliation{%
    \institution{Google DeepMind}
    \country{Japan}
 }
\email{yujintang@google.com}


\renewcommand{\shortauthors}{Lange et al.}

\begin{abstract}
Large Transformer models are capable of implementing a plethora of so-called in-context learning algorithms.
These include gradient descent, classification, sequence completion, transformation, and improvement. 
In this work, we investigate whether large language models (LLMs), which never explicitly encountered the task of black-box optimization, are in principle capable of implementing evolutionary optimization algorithms.
While previous works have solely focused on language-based task specification, we move forward and focus on the zero-shot application of LLMs to black-box optimization. We introduce a novel prompting strategy, consisting of least-to-most sorting of discretized population members and querying the LLM to propose an improvement to the mean statistic, i.e. perform a type of black-box recombination operation.
Empirically, we find that our setup allows the user to obtain an LLM-based evolution strategy, which we call `EvoLLM', that robustly outperforms baseline algorithms such as random search and Gaussian Hill Climbing on synthetic BBOB functions as well as small neuroevolution tasks.
Hence, LLMs can act as `plug-in' in-context recombination operators.
We provide several comparative studies of the LLM's model size, prompt strategy, and context construction.
Finally, we show that one can flexibly improve EvoLLM's performance by providing teacher algorithm information via instruction fine-tuning on previously collected teacher optimization trajectories.
\end{abstract}

%
%



\begin{teaserfigure}
  \centering
  \includegraphics[width=0.995\textwidth]{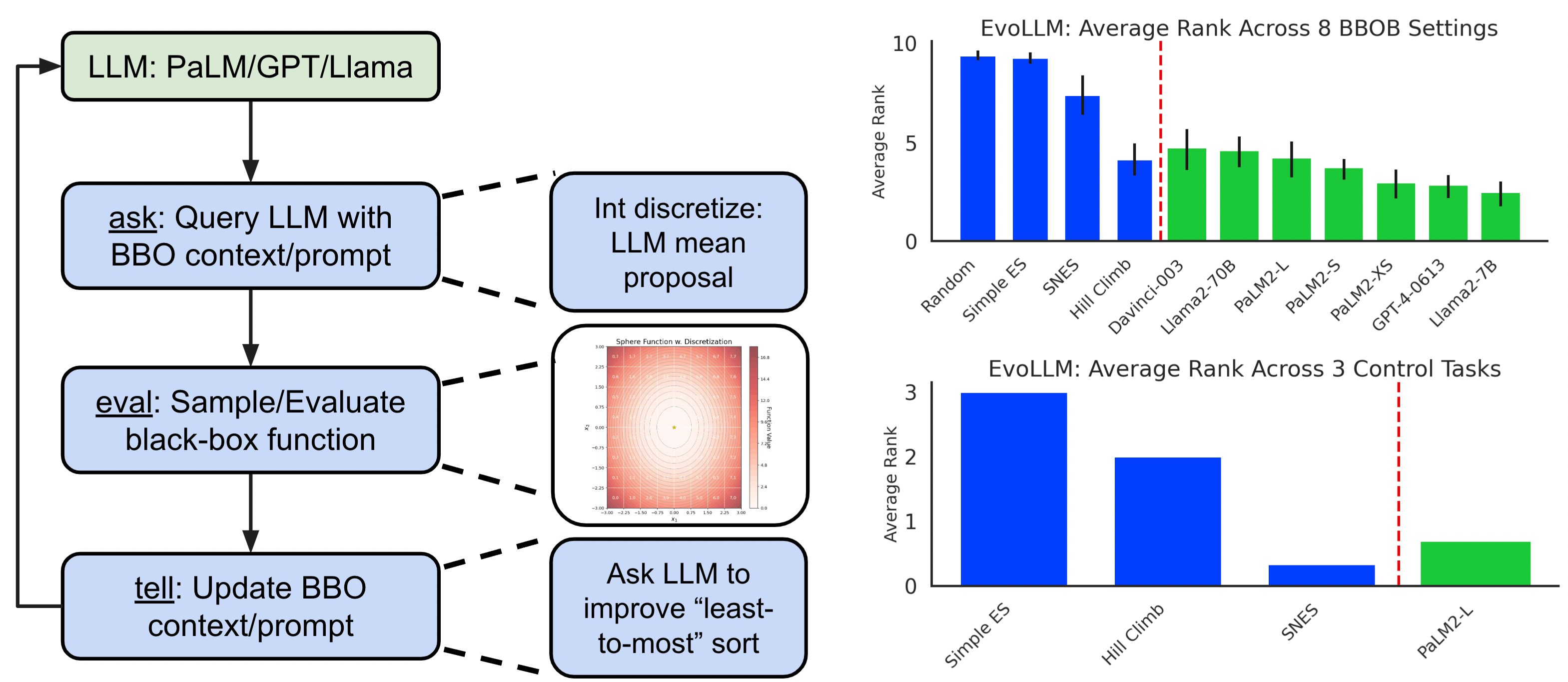}
  \caption{Left. Overview of EvoLLM procedure. An LLM proposes an ES search distribution update using a discretized search space \& solutions sorted by their performance (least-to-most). To combat the context length growth in the number of dimensions, we can split the search dimensions into blocks and perform LLM batch queries. Right: Aggregated results across 8 BBOB \citep{hansen2010real} settings, and 3 neuroevolution control problems. The results are averaged over ten and five independent runs, respectively. LLM-based Evolution Strategies (green) outperform traditional baselines (blue). 
  }
  \label{fig:conceptual}
\end{teaserfigure}

\maketitle

\newpage
\section{Introduction}

\textbf{Motivation}. Recently, it has been demonstrated that language models trained on large text corpora are capable of impressive in-context learning~\cite{brown2020language}. For example, given a prompt of pattern demonstrations, LLMs can infer the underlying generating rules and even propose sequence improvements~\cite{mirchandani2023large}.
Unlike fine-tuning this property notably emerges with frozen model weights and only requires online context information. Interestingly, this in-context improvement property appears to be broadly applicable to abstract pattern sequences: In the context of in-silico evolution, LLMs can out-of-the-box act as code-level mutation~\cite{chen2023evoprompting}, cross-over operations~\cite{meyerson2023language} or be embedded into the context of genetic programming~\cite{lehman2023evolution}. 
%
To what degree can LLMs, trained on text, be generalized to behave like an optimization algorithm? Is it possible to have an LLM train the weights of a neural network as in evolutionary strategies (ES)? \\
\textbf{Approach}. Here, we leverage the recent advances in the understanding of LLMs as ‘general pattern machines’~\cite{mirchandani2023large} to construct a prompt strategy, which turns an LLM into a recombination operator:  A purely text-trained LLM processes fitness-sorted sequences of function evaluations and their corresponding candidate solutions. Afterwards, we can simply `ask' the LLM to propose the next mean statistic to sample from (see \cref{fig:conceptual}, left). We term the resulting LLM-based Evolution Strategy \textit{EvoLLM}. Concretely, we also propose to adapt in our approach integer-based, search space discretization, least-to-most prompting~\cite{zhou2022least} and decision transformer-style~\cite{chen2021decision} fitness improvement queries.\\
\textbf{Results}. Our experiments demonstrate that our proposed prompting strategy is sufficient to induce the LLM to conduct a recombination operation. The resulting EvoLLM successfully performs black-box optimization (BBO) on both synthetic BBOB~\cite{hansen2010real} test functions and small neuroevolution tasks~\cite{brockman2016openai, gymnax2022github} with various search space dimensions and evaluation budgets (see \cref{fig:conceptual}, right). Furthermore, we provide rigorous ablation to the prompt design and provide general construction recommendations promoting better optimization behavior. 
Interestingly, we observe that among our selection of LLMs, there is a general trend indicating that smaller LLM models tend to outperform larger ones.
We also show that choosing a sufficient solution representation is critical for in-context BBO powered by LLMs. 
Finally, LLMs are capable of leveraging additional information provided by supporting teacher algorithms.\\
\textbf{Contributions}. Our contributions are summarized as follows:

\begin{enumerate}
\item We introduce a general prompt approach that induces LLMs to act as ES. The prompt is composed of a discretized solution candidate representation, performance-based least-to-most sorting and a fitness improvement query (\cref{fig:conceptual}).
\item We use a set of diverse LLMs and establish that language models can robustly perform zero-shot optimization on classic BBO and small neural network control tasks.
\item We investigate various prompting strategies and show that a discretized solution space representation greatly outperforms common natural language-based instructions. Furthermore, the EvoLLM approach is largely robust to the choice of search space discretization and context information.
\item We show that EvoLLM's performance can be improved by fine-tuning the base LLM model on BBO trajectories generated by teacher algorithms.
\end{enumerate}

\section{Related Work}

\textbf{In-Context Learning with Transformers}. Various recent efforts have investigated the possibility of training large models from scratch to perform in-context learning. These methods rely on the impressive sequence modeling capabilities of the Transformer architecture~\cite{vaswani2017attention} to model long-range dependencies. For example \citet{kirsch2022general} show that Transformers can learn supervised in-context image classification algorithms. \citet{laskin2022context} investigated in-context reinforcement learning and \citet{lu2023structured} considered state space models for large-scale meta-reinforcement learning. \\
%
\textbf{In-Context BBO with Autoregressive Models}. \citet{chen2017learning} trained RNN-based BBO algorithms using privileged access to gradient computations of the fitness function at training time. Furthermore, \citet{dery2022multi, chen2022towards} used large pre-collected datasets to directly train a T5 Encoder-Decoder architecture~\cite{raffel2020exploring} for BBO, called OptFormer. \citet{krishnamoorthy2022generative} later on investigated the usage of generating offline data for training autoregressive BBO models. All of these approaches trained large sequence models leveraging large programmatically generated or augmented task spaces to induce in-context learning across long timescales.
While this line of works is close to our proposed method, ours differs in that we use text-trained large models and investigate their ability to perform BBO without explicitly being trained on such tasks.\\
\textbf{In-Context Learning with LLMs}. LLMs are capable of few-shot learning given little text examples in their prompt~\cite{brown2020language}. Various prompting strategies and automation methods have subsequently been developed to improve task-specific performance. E.g. these include least-to-most sorting ~\cite{zhou2022least}, chain-of-thought prompting~\cite{wei2022chain} or self-consistency~\cite{wang2022self}. More recently, \citet{mirchandani2023large} have shown that these LLM capabilities can also be elicited outside of the text-based prompting context. Indeed LLMs appear to be able to reason about abstract sequences of integers or even ASCII codes.\\
%
\textbf{LLMs for Optimization}. Related to our work, \citet{yang2023large, zhang2023using, nie2023importance} show that LLMs can be turned into text-based optimizers. Our work extends this line of work and shows that they can also be transformed into a recombination operator for ES and are capable of optimizing small neural networks. In the context of computational evolution, LLMs can out-of-the-box act as code-level or algorithm mutation~\cite{chen2023evoprompting, liu2023algorithm}, cross-over operations~\cite{meyerson2023language}, be embedded into the context of genetic programming~\cite{lehman2023evolution} or evolve prompt strategies~\cite{guo2023connecting}. 
While preliminary work~\cite{liu2023large, liu2023large_multi} has started to investigate LLMs for Evolutionary Optimization, our work is the first to consider LLMs for ES, compares different base LLM models and various prompt construction approaches.\\
\textbf{Learned Black-Box Optimization}. ES-update rules are inherently set of operations, i.e. the order of the population members within a generation should not affect the performed distribution change. Self-attention provides a natural inductive bias for such an operation. Previously, \citet{lange2023discovering_es, lange2023discovering_ga} constructed ES and GA algorithms, which used self- and cross-attention to process the information within a single generation. The attention parameters are meta-evolved on a small task distribution of BBO problems. As a comparison, our proposed method does not meta-train a new BBO algorithm but instead asks whether text-trained LLMs are capable of performing optimization with discretized prompt information provided.

\section{Background}
\textbf{Black-Box Optimization (BBO)}. Given a function $f(\mathbf{x}): \mathbb{R}^D \to \mathbb{R}$ with unknown functional form, i.e. we cannot compute its derivative (or it is not well behaved or empirically infeasible to compute), BBO seeks to find its global optimum using only function evaluations, without derivatives:
%
\begin{align*}
    \min_\mathbf{x} f(\mathbf{x}), \text{s.t.} \ \mathbf{x}_d \in [l_d, u_d] \subset [-\infty, \infty], \forall d=1,\dots,D.
\end{align*}
%
Throughout, we leverage a set of synthetic benchmark functions (BBOB~\cite{hansen2010real} and classic control tasks~\cite{brockman2016openai} to evaluate BBOs.\\
\textbf{Evolution Strategies}. Evolutionary Optimizers (EO) are a set of BBO algorithms inspired by mechanisms of biological evolution. Roughly speaking, EO algorithms can be grouped into Evolution Strategies~\cite{rechenberg1978evolutionsstrategien} and Genetic Algorithms, where the former focus on mutating real-valued parameters and often use self-adaptation, while the latter typically work with a population of binary-coded solutions and rely on crossover and mutation operators.
Here, we focus on isotropic Gaussian ES. Given a population size $N$ and a search distribution $\boldsymbol{\mu} \in \mathbb{R}^D, \Sigma = \sigma \mathbf{1}_{D \times D} \in \mathbb{R}^{D \times D}$, ES sample a population of candidate solutions $X = [x_1, \dots, x_N] \in \mathbb{R}^{N \times D}$ at each generation. Afterwards, the performance (or fitness) of each candidate is evaluated on the task of interest and one obtains fitness scores $F=[f_1, \dots, f_N] \in \mathbb{R}^N$ (we denote $[f(x_1), \dots, f(x_N)]$ as $[f_1, \dots, f_N]$ for brevity). The search distribution is then updated to increase the likelihood of sampling well-performing solutions, $\boldsymbol{\mu}', \boldsymbol{\sigma}' \leftarrow \texttt{UPDATE}(\boldsymbol{\mu}, \boldsymbol{\sigma}, X, F, H)$, where $H$ denotes a set of summary statistics constructed from the search history. There exist various types of ES including estimation-of-distribution~\cite{hansen2006cma}, natural~\cite{schaul2011high, wierstra2014natural} and finite-difference-based ES~\cite{salimans2017evolution}. In this work, we investigate whether an LLM can represent a $\texttt{UPDATE}$ operator using a context string constructed from $H$.\\
\textbf{Transformer-Based Language Models}. The Transformer~\cite{vaswani2017attention} stacks blocks of multi-head self-attention (MHSA) operations, feedforward MLP transformation, dropout, and layer normalization.  At its core self-attention is a set operation that projects an input matrix $X \in \mathbb{R}^{T \times D}$ onto $D_K$-dimensional vectors $Q, K, V \in \mathbb{R}^{T \times D_K}$ called queries, keys, and values, respectively:

\vspace{-0.25cm}
\begin{align*}
    \mathrm{Attention}(X) &= \mathrm{softmax}\left(QK^T/\sqrt{D_K}\right) V \\
    &= \mathrm{softmax}\left(X W_Q \left(X W_K\right)^T/\sqrt{D_K}\right) X W_V.
\end{align*}

Permuting the rows of $X$ will apply the same permutation to $\text{Attention}(X)$ \cite[e.g., ][]{lee2019set, tang2021sensory, kossen2021self}. MHSA has quadratic complexity with respect to the context length $T$. In this work, we leverage large Transformer models trained on text data. Afterwards, we evaluate their ability to implement evolutionary improvement operations when presented with BBO trajectory information, i.e. discretized solution candidates and their fitness or rank. Hence, we ask whether in-context learning can resemble an efficient $\texttt{UPDATE}$ operation.\\
\textbf{Prompt Engineering \& Vocabulary Tokenization}. LLMs can behave sensitively concerning the specific context construction. This has led to terms such as `prompt engineering'. Throughout our investigations, we consider various prompt construction approaches to provide ablative insights into the underlying LLM mechanisms. More specifically and motivated by computational resource considerations, we conduct coordinate-wise evaluations in order to extract a well-performing approach. 
There exist several automated prompt tuning approaches including gradient-based soft prompt optimization~\cite{lester2021power}, which requires access to LLM weights. Random search-based prompt optimization~\cite{lu2023prompt}
and evolutionary optimization of prompts~\cite{li2023spell, guo2023connecting, fernando2023promptbreeder} have recently shown promising result.

Language models pre-process raw language strings into vector representations using so-called tokenizers. These tokenizers perform a type of compression based on natural language co-statistics. Naturally, raw high-precision floating point numbers used in BBO tend to be underrepresented. We therefore, propose a different integer-based representation approach.

\section{Turning LLMs into ES Algorithms}
\label{sec:evollm_prompt}
In the following, we outline a prompt strategy that enables LLM-based optimization in the form of an Evolution Strategy. More specifically, we establish a prompt design space used to construct an LLM query, which makes the LLM represent the $\texttt{UPDATE}$ operator for an ES. Afterwards, we show how this procedure can be incorporated into the simple `ask-evaluate-tell` API common to many popular black-box optimization algorithms~\cite{lange2022evosax}.\\
\begin{figure*}[h]
    \centering
    \includegraphics[width=0.9\textwidth]{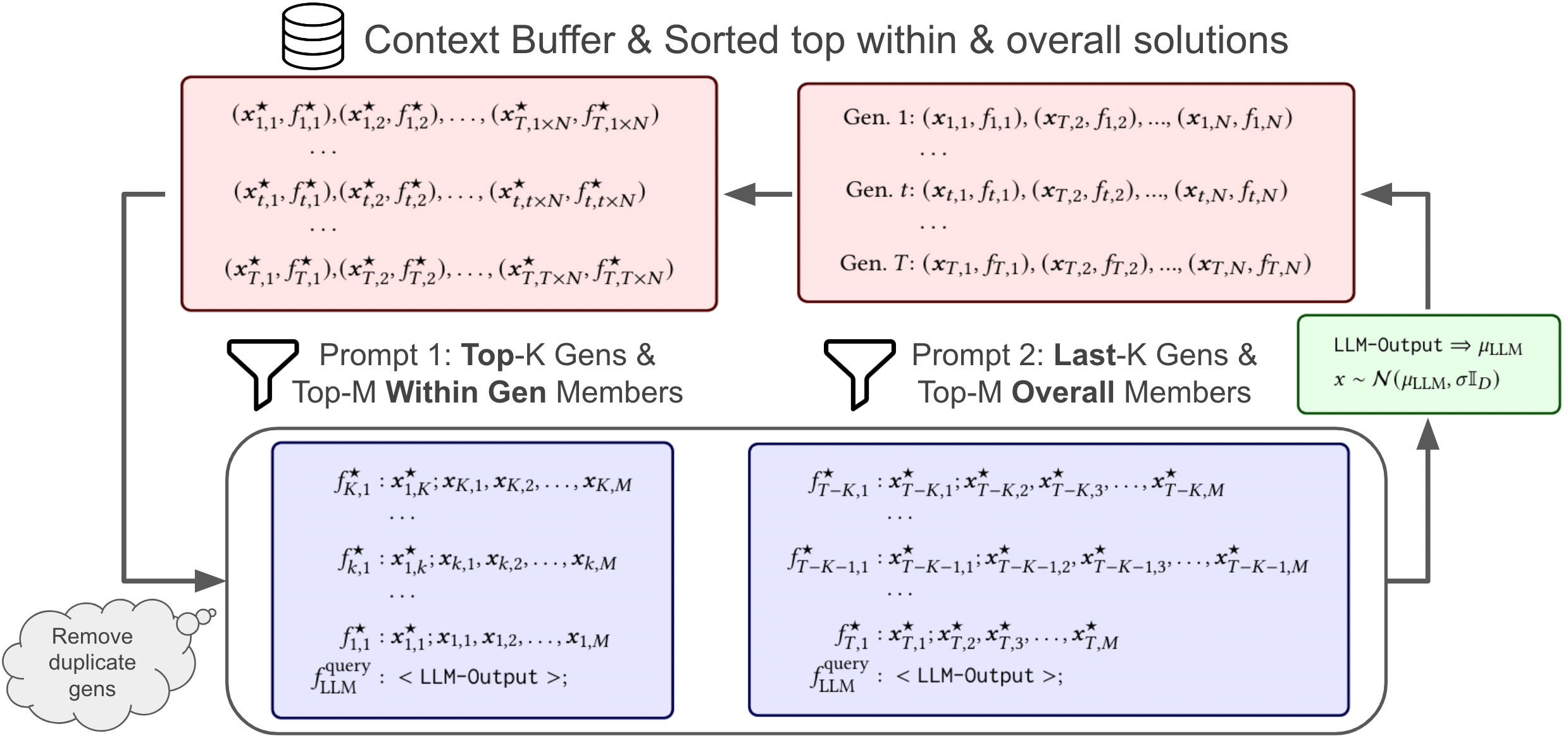}
    \caption{EvoLLM prompt design space \& API. We track all solution evaluations and their performance in a context buffer. The buffer is used to construct query prompts for the LLM. After parsing the LLM output and sampling, we evaluate the resulting population and add the new information to the buffer. We provide an example of the generated prompts in the appendix.}
    \label{fig:design_space}
\end{figure*}
\textbf{High-Level EvoLLM Prompt Design Space}. We follow the paradigm outlined in \citet{mirchandani2023large} and construct a LLM prompt by representing the solution candidates as integers resulting from a disretized search space with a pre-specified resolution (\cref{fig:conceptual}). Note, that we don't use raw floating point numbers due to the LLM tokenizer potentially returning different numbers of tokens per individual number. This can hinder the LLM from inferring improvement sequences. 
%
First, we sort the set of previous population evaluations $H=\{X_g, F_g\}_{g=1}^G$ by their fitness within and across generations. Afterwards, we select the top-$K$ performing generations and top-$M$ solutions within each generation. We let the LLM propose the next mean for a desired fitness level~\cite{chen2021decision}, $f^{\text{query}}_{\text{LLM}}$:

\begin{empheq}[box=\mymath]{align*}
    f^{\star}_{K, 1}: & \ \boldsymbol{x}^{\star}_{1, K}; \boldsymbol{x}_{K, 1}, \boldsymbol{x}_{K, 2}, \dots, \boldsymbol{x}_{K, M} \\
    & \dots \\
    f^{\star}_{k, 1}: & \ \boldsymbol{x}^{\star}_{1, k}; \boldsymbol{x}_{k, 1}, \boldsymbol{x}_{k, 2}, \dots, \boldsymbol{x}_{k, M} \\
    & \dots \\
    f^{\star}_{1, 1}: & \ \boldsymbol{x}^{\star}_{1, 1}; \boldsymbol{x}_{1, 1}, \boldsymbol{x}_{1, 2}, \dots, \boldsymbol{x}_{1, M} \\
    f^{\text{query}}_{\text{LLM}}: & \ <\texttt{LLM-Output}>;
\end{empheq}
where $x^\star_k, f^\star_k$  denotes the best-performing solution and its fitness up to generation $k$. We observed that LLMs robustly follow the pattern outlined in the prompt above and continue the string format by outputting a new mean $x_{LLM}$ with the delimiter $;$. Afterwards, we use the proposed mean to sample a new set of candidates, update the context statistics $H$ and iterate the process. The LLM can thereby in-context adapt to accumulated search information and implement a novel type of recombination.\\
\textbf{Detailed EvoLLM API}. At each generation, ES samples a set of candidates, evaluates the population on the task at hand and afterward updates the search distribution. Here, we let the LLM perform the search update given text-context information $H$. The general procedure consists of the following steps (\cref{fig:design_space}):


\begin{enumerate}
    \item \textbf{Context Buffer Warm-Up/Seeding}. We use standard BBO algorithm (here: random search) to fill up a context data buffer with an initial set of evaluations.
    \item \textbf{Discretize \& Augment Context Buffer}. We represent the solutions as discretized integers with a chosen resolution and track the evaluation candidates and their fitness scores.
\end{enumerate}
Given the initial buffer with discretized solutions, we construct a string representation of the previous evaluations. The $K$ generations and corresponding $M$ candidates are selected and sorted as follows:

\begin{enumerate}[resume]
    \item \textbf{Select \& Sort Context Generations}.
    \begin{enumerate}
        \item \textbf{`Random'}. The simplest option is to select previous generations from the buffer uniformly at random. 
        \item \textbf{`Last'}. Alternatively, we select the most recent $K$ generations evaluated on the problem.
        \item \textbf{`Best'}. Finally, we considered the generations, which led to the best-performing candidate solutions seen so far.
    \end{enumerate}
    \item \textbf{Select \& Sort Context Candidates}.
    \begin{enumerate}
        \item \textbf{`Random'}. From the $K$ selected generations we again select $M < N$ random population members.
        \item \textbf{`Best-Within-Generation'}. Alternatively, we select the best candidates evaluated within a given generation.
        \item \textbf{`Best-Up-To-Generation'}. Finally, we consider the entire evaluation history and provide the top-$M$ candidate solutions evaluated up to the $k$-th generation. 
    \end{enumerate}
\end{enumerate}
%
Note, that the exact ordering of the generations and candidates can affect the ease with which the LLM may infer improving directions (e.g. momentum). Each generation of candidates is separated with a line break and each member with a $,$. Finally, we add a fitness improvement query (ca. 2 times the best previously seen fitness value). After querying the LLM, we parse the LLM-proposed mean update back into a floating point representation:

\begin{enumerate}[resume]
    \item \textbf{Query LLM for Search Improvement}. We construct the prompt repeatedly at each generation, sample a temperature parameter and query the LLM. We pass the returned integer output back into the mean for the next generation. Occasionally, the parsing may fail. In this case, we use a back-up strategy of sampling around the previous best evaluation.
    \item \textbf{Sample \& Evaluate New Candidates}. We perturb the mean with isotropic noise and evaluate all population members. Afterwards, we add the information to the context buffer.
\end{enumerate}
\begin{figure}[h]
    \centering
    \includegraphics[width=0.3\textwidth]{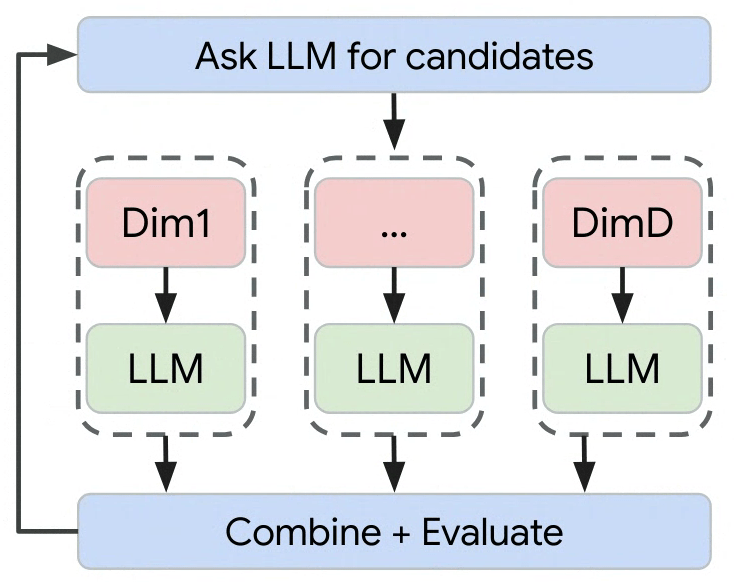}
    \caption{Dimension-batched Querying of an LLM. As the search space dimensionality grows, the context length can exceed the feasibilities of the LLM. We split the solution space into blocks and perform multiple LLM queries per update.}
    \label{fig:method}
\end{figure}
%
\begin{figure*}[h]
    \centering
    \includegraphics[width=0.975\textwidth]{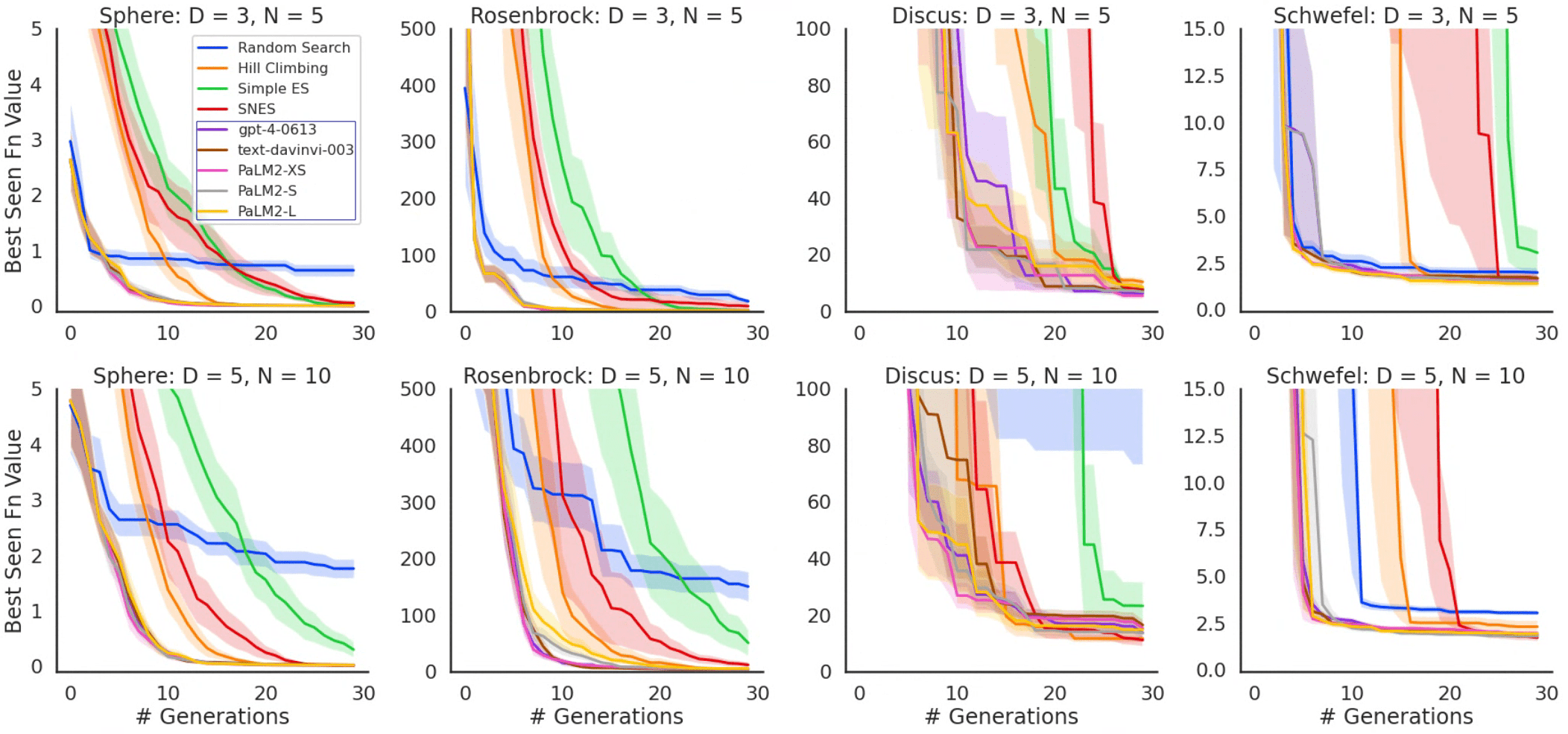}
    \caption{EvoLLM performance (\underline{lower is better}) on BBOB~\cite{hansen2010real} functions with single LLM query. We compare different LLM base models (marked in the lower box in the legend) and find that the behavior of EvoLLM is robust to the exact choice of LLM. The results are averaged across 10 independent runs.}
    \label{fig:results_bbob}
\end{figure*}
\textbf{Scaling EvoLLMs to Larger Search Spaces}. The context length of the used prompt quickly increases with the number of dimensions. Many LLMs have been trained with a context length with limits their applicability to longer horizons. We found that once the context becomes too long, the LLM may output non-informative information. This in turn implies that either the number of context generations or the number of considered context population members per generation would have to be reduced to allow for large search spaces. To avoid this limitation we use (block-)independent LLM queries for batches of dimensions. More specifically, we group a set of dimensions that fits into the context of the LLM and perform multiple LLM queries per generation (\cref{fig:method}). Hence, we trade off an increased LLM inference time with scalability to a larger number of search dimensions. 
In the limit, each LLM call processes a single dimension $d$. 
Note that as the capabilities of LLMs to model longer-range dependencies increase, EvoLLM will likely benefit from such advances.


\section{LLMs are Zero-Shot Evolution Strategies}

\subsection{Evaluation on Synthetic BBO Functions}

We start by evaluating the performance of the EvoLLM prompt design on various BBOB~\cite{hansen2010real} tasks with different numbers of search dimensions and population sizes. We use the $K=5$ last generations and $M = 5$ best-seen evaluations throughout the optimization trajectory to construct the context string. We use a set of 4 warm-up generations using random search to seed the context buffer $H$. Afterwards, we use the LLM proposed mean update to perform BBO and use a fixed perturbation strength, $\sigma=0.2$. The default prompt construction settings are summarized in \cref{table:context_settings}.

\begin{table}
\begin{tabular}{ |p{3cm}|p{3cm}| }
\hline\hline
Setting Name & Setting Choice \\
\hline
Context Generations & $K=5$\\
Context Members & $M=5$ \\
Generation Selection & 'last' \\
Candidate Selection & 'best-across' \\
Generation Sorting & 'improving' \\
Candidate Sorting & 'improving'\\
Improvement Indicator & \checkmark\\
Uniqueness Filtering & False \\
Improvement Querying & True \\
Warm-up Generations & 4 \\
Warm-up Strategy & Random Search\\
\hline
\end{tabular}
\caption{EvoLLM Context Construction Settings. 
}
\label{table:context_settings}
\end{table}
To assess the performance of EvoLLM across various settings, we consider four different tasks, which consist of the separable Sphere, moderate-condition number Rosenbrock, the high-condition number Discus, and the multi-modal Schwefel function.
\cref{fig:results_bbob} shows that the LLM-based ES can outperform random search and Gaussian Hill Climbing on these BBOB functions~\cite{hansen2010real} with different search dimensions ($D \in \{3, 5\}$) and population sizes ($N \in \{5, 10\}$). 
On many of the considered tasks, the LLM-based Strategy is even capable of outperforming diagonal covariance ES such as SNES~\cite{schaul2011high}. Furthermore, EvoLLM outperforms all baselines across all tasks for small budget settings, i.e. less than 10 generations.
\begin{figure}[h!]
    \centering
    \includegraphics[width=0.475\textwidth]{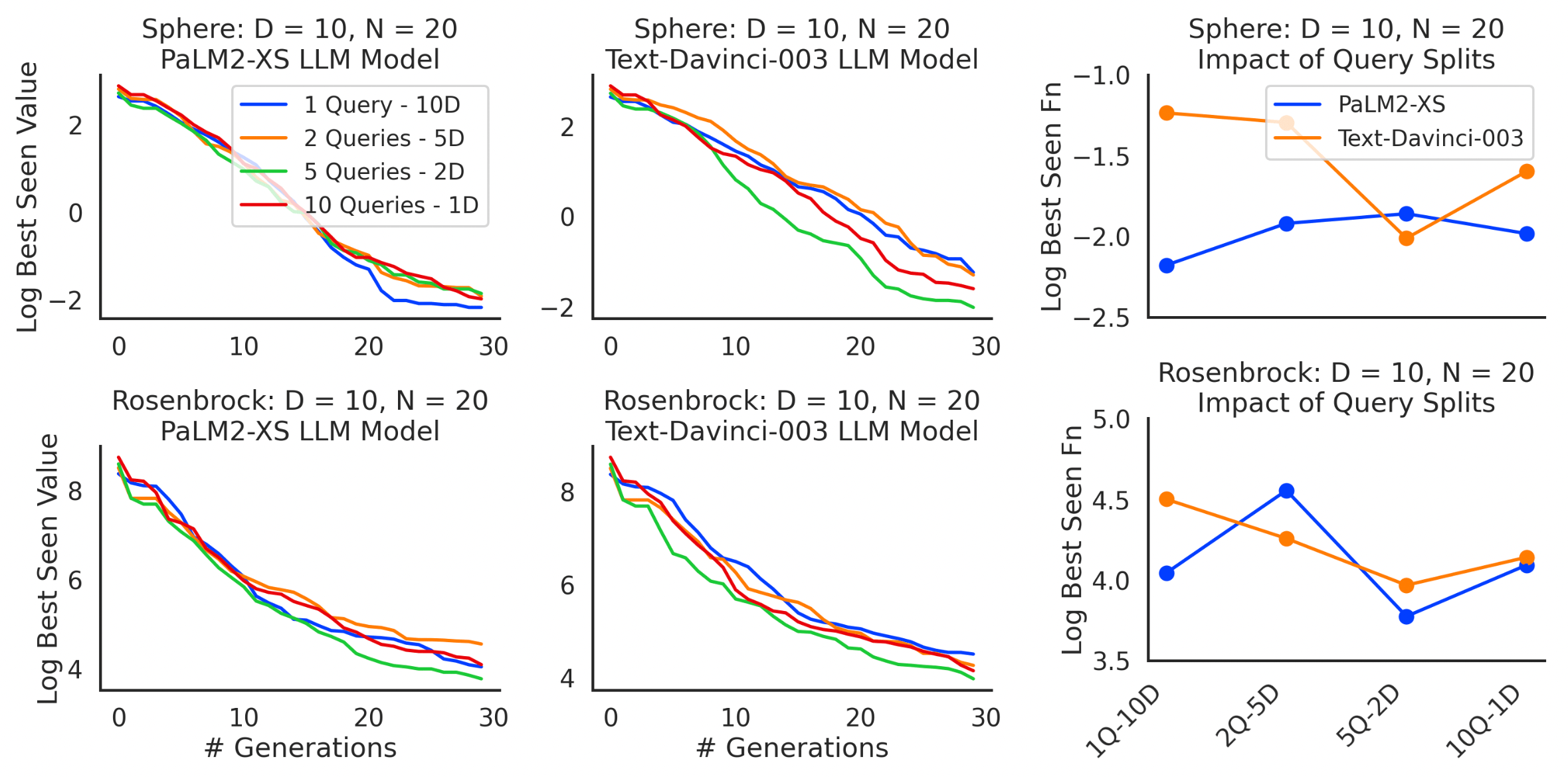}
    \caption{EvoLLM performance (\underline{lower is better}) on BBOB~\cite{hansen2010real} functions with multi-dimensional LLM query splits. We consider text-davinci-003 and PaLM2-XS as base LLM models and find that performance does not degrade when using splits. Top: 10-dimensional Sphere problem. Bottom: 10-dimensional Rosenbrock problem. Averaged results over 5 independent runs.}
    \label{fig:results_multidim}
\end{figure}
We consider three different classes of differently sized pre-trained LLM models including three PaLM2 models~\cite{palm2}, OpenAI models~\cite{OpenAI_GPT4_2023}, and the open-source available Llama2~\cite{touvron2023llama} models.
We observe (\cref{fig:conceptual}, right) that the LLM model size inversely affects the downstream performance of EvoLLM. Larger models (PaLM-L, Llama2-70B) tend to perform worse than smaller models (PaLM-XS, Llama2-7B), negating a scaling law trend. Interestingly, GPT-4 tends to perform similarly to the smaller models potentially pointing towards the size of the individual mixture of expert models.\\
%
Next, we scaled EvoLLMs to larger search spaces ($D = 10$) using batched LLM queries for blocks of parameters. More specifically, we considered different numbers of queries and dimension groupings, e.g. 5 times 2-dimensional versus 2 times 5-dimensional LLM queries. Thereby, the resulting optimizer has to perform blocked separable optimization and can not interpolate information from other potentially correlated groups of dimensions.
\cref{fig:results_multidim} indicates that the performance of the EvoLLM does not significantly decline as we optimize the parameters in groups. Interestingly, this observation holds for both the separable quadratic fitness function as well as the non-separable Rosenbrock function. Furthermore, it it robust for two different model classes, PaLM2-XS and text-davinci-003.

\begin{figure}[h!]
    \centering
    \includegraphics[width=0.475\textwidth]{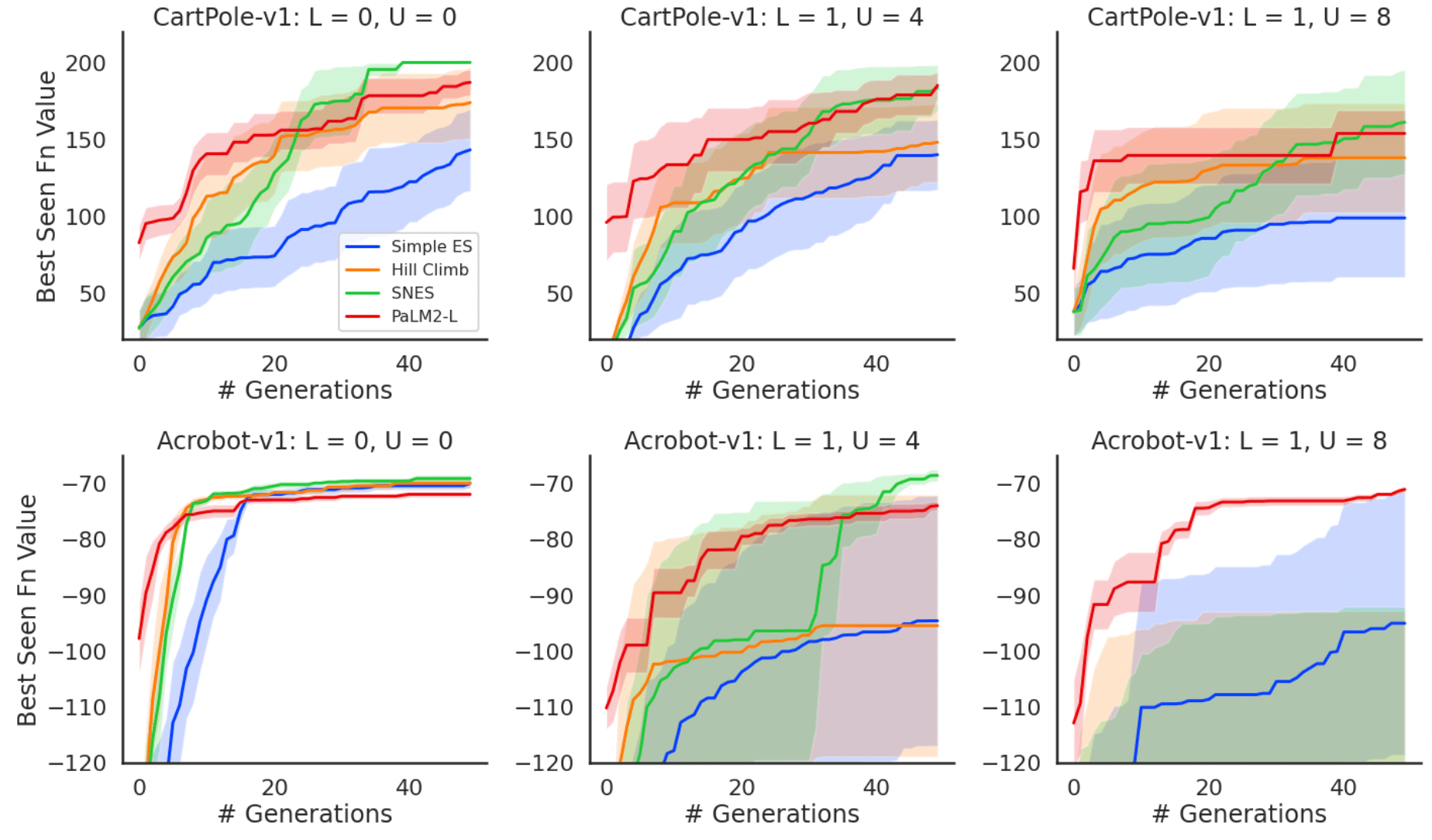}
    \caption{EvoLLM performance (\underline{higher is better})  on CartPole \& Acrobot~\cite{brockman2016openai, gymnax2022github} control task with different neural network architectures. LLM-based ES can optimize small networks and even outperform baselines in the small evaluation budget regime. Averaged results over 5 independent runs.}
    \label{fig:results_neuroevolution}
\end{figure}

\begin{figure*}[h!]
    \centering
    \includegraphics[width=0.85\textwidth]{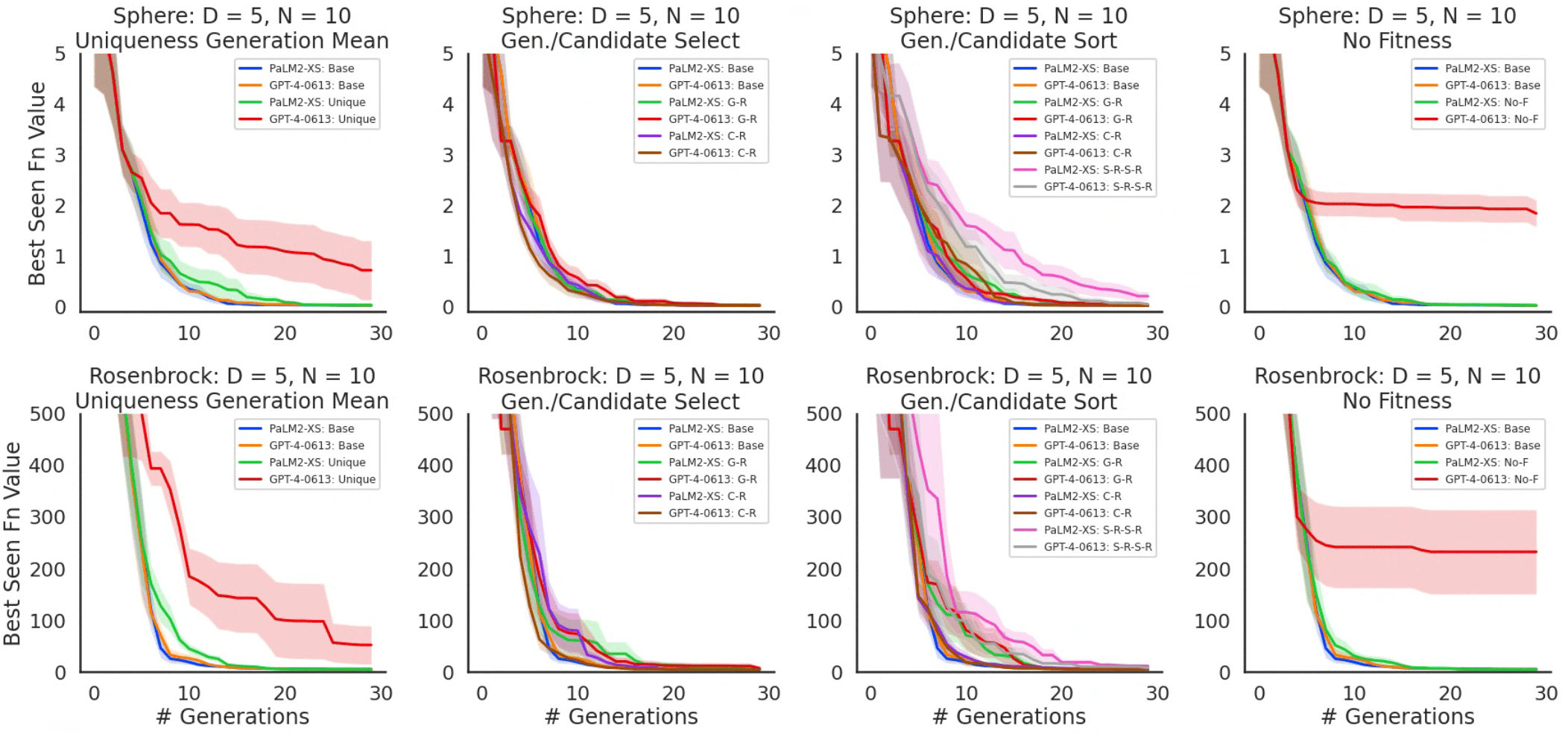}
    \caption{Different Prompt Constructions on two different BBOB problems. Left: Impact of generation uniqueness filtering. Middle: Impact of selection and sorting of generations and candidates. Right: Impact of providing fitness information and improvement query. The EvoLLM prompt is largely robust to all individual choices, but the performance drops if we do not include the fitness information or filter for improving generation sequences.  Averaged results over 5 independent runs.}
    \label{fig:results_prompt}
\end{figure*}

\subsection{Evaluation on Neuroevolution Tasks}

The previous BBO results indicate that LLM-based ES are capable of optimizing various classic functions with different characteristics (conditioning, single/multi-modal optima structure, etc.). However recent work has shown that benchmarking on such functions can be of limited relevance for machine learning tasks~\cite{lange2023neuroevobench}. Therefore, we wondered whether EvoLLMs can also be applied to neuroevolution tasks. If this is indeed the case, LLMs may be a viable future option for large-scale autonomous optimization which potentially can include various text-based information.
We consider the CartPole-v1 and Acrobot-v1 discrete control tasks using a single hidden layer MLP policy. In this case, the EvoLLM has to evolve the parameters of the feedforward network used to output the agent's action at every episode timestep. The considered networks contain between 16 and 40 weights and biases to be optimized. We again batch the parameters into groups and optimize the neural network parameters using 32 population members and 8 stochastic rollouts for fitness evaluation.
\cref{fig:results_neuroevolution} provides evidence that EvoLLM can indeed evolve such neural network-parametrized policies for both considered tasks. More specifically, it is again capable of even outperforming competitive baselines in the small budget regime.
We note that optimization becomes more challenging as the number of optimized parameters increases.
This result is of importance because it provides an intriguing perspective on LLM-based agents: In principle, they can optimize neural network artifacts using a gradient-free implicit optimization procedure implemented by their internal activations. 

\section{EvoLLM Ablation Studies}

After having established that LLMs are capable of acting as improvement operators for ES, we next investigate the importance of the prompt design, discretization resolution, and context length.

\subsection{Prompt Strategy Ablations}

We consider the following 4 different prompt design decisions:
\begin{enumerate}
\item \textbf{Uniqueness of mean for the selected generations}: We either filter the context generations by the uniqueness of their mean in the sequence or not (\cref{fig:results_prompt}, left).
\item \textbf{Selection criteria for context generations/candidates}: We compare the base selection setting (\cref{table:context_settings}) with randomly selected generations ('G-R') and members ('C-R').
\item \textbf{Sorting criteria for selected generations/candidates}: We compare the base sorting setting (\cref{table:context_settings}) with randomly sorted generations ('G-R') and members ('C-R').
\item \textbf{Fitness information provision \& improvement request}: We either add the desired fitness query or not.
\end{enumerate}
\cref{fig:results_prompt} considers two different BBOB settings (Sphere, Rosenbrock) and two different base LLM models (GPT-4, PaLM-XS). It is not beneficial to filter the selected generations by mean uniqueness. Furthermore, EvoLLM remains largely robust concerning the selection of context candidates and members. The sorting criterion, on the other hand, has strong effects on the downstream performance. Sorting generations and members randomly decreases performance substantially. Furthermore, not providing fitness information decreases GPT-4's performance. This suggests that different LLM models can behave differently in the context of BBO.

\subsection{Raw text vs. discretized representation}

One key aspect of the EvoLLM is our usage of discretized integer representations instead of raw floating point number strings. In preliminary experiments, we observed that the vocabulary tokenizer of LLMs struggles with representing high-precision numbers. Different numbers may result in different numbers of tokens (\cref{fig:supp_tokenizer}), which in turn makes it challenging for the LLM to infer improvements. Most common tokenizers (e.g. SentencePiece) represent integer numbers up to 1000 as a single token. We therefore chose to translate raw solutions into integer representations and map the LLM output back into the corresponding floating point.
To illustrate the impact of this choice, we compare the performance of our EvoLLM with the raw text-based prompting strategy outlined in \citet{zhang2023using}. Using the same LLM backend model we repeatedly observe that the text-based approach is not capable of 'zooming into' optima and instead performs too large perturbations to progress after initial improvements.

\begin{figure}[h]
    \centering
    \includegraphics[width=0.35\textwidth]{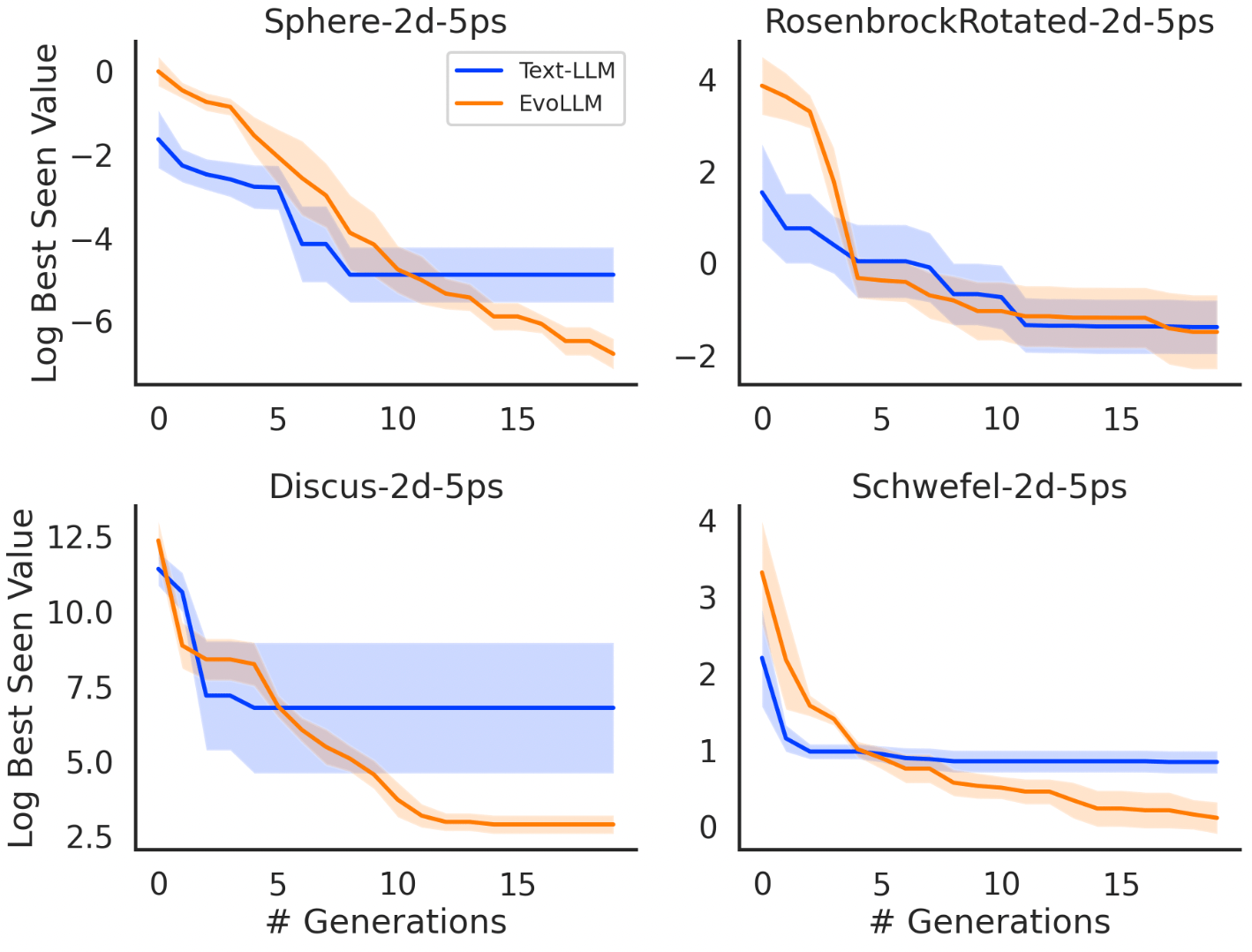}
    \caption{EvoLLM versus text-based optimization prompting performance. Our proposed discretized solution representation is capable of improvements even for longer optimization trajectories. Text-based prompting, on the other hand, quickly saturates. Averaged results over 5 independent runs.}
    \label{fig:results_text_prompt}
\end{figure}

\subsection{Impact of Resolution \& Context Length}
Next, we considered how the chosen discretization resolution impacts the EvoLLM performance. Intuitively, a resolution that is too coarse will not allow the optimizer to discover very narrow optima basins, while a too-high resolution will hurt the LLM's ability to infer improvements from the context. We used the template outlined in \cref{table:context_settings} and discretized an optimization range from -3.0 to 3.0 into $\{50, 100, 1000, 10000\}$ bins. In \cref{fig:results_resolution} we indeed observe the hypothesized behavior for two different PaLM2 models: As we increase the resolution up to 1000, performance improves. For 10000, on the other hand, it decreases again. For the Rosenbrock problem, on the other hand, this trend is not clear. This may be due to the optimizer not finding the global optimum but preemptively converging in a broader local optimum basin. This indicates that the optimal resolution parameter is indeed problem-specific. 

\begin{figure*}
    \centering
    \includegraphics[width=0.85\textwidth]{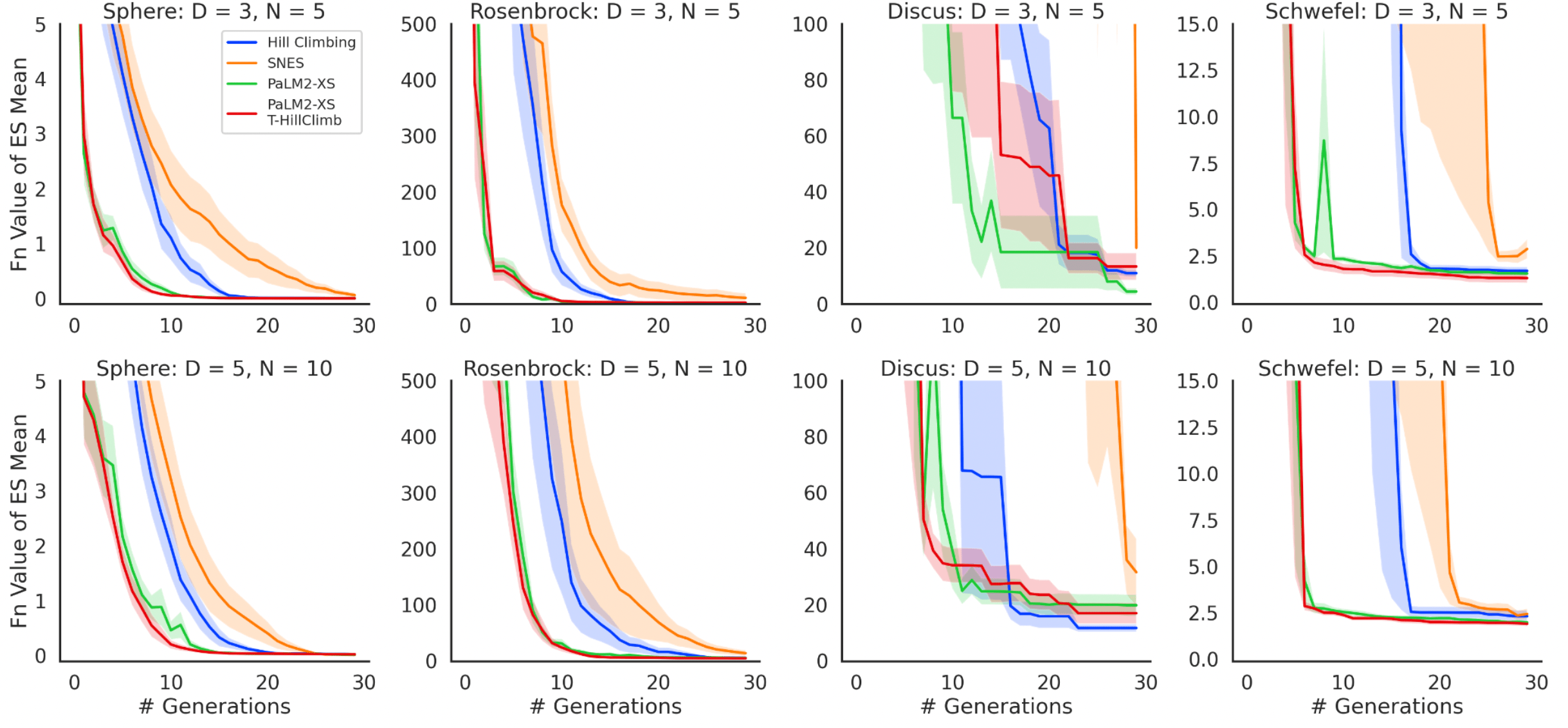}
    \caption{Instruction fine-tuning using the Hill-Climbing algorithm improves EvoLLM's performance on unseen BBOB tasks (\underline{lower is better}). Results are averaged over 5 independent runs.}
    \label{fig:results_finetune}
\end{figure*}

Finally, we wondered how much context information is required for the LLM to infer beneficial updates to the mean statistic. In \cref{fig:results_context_info} we considered different amounts of context generations and members. We find that even with extremely limited information, the EvoLLM can make improving updates to the mean. Often more context generations appear to be beneficial compared to more context members. This may be due to the clear improvements in the sequence, which makes it easier for the LLM to infer continuation patterns such as momentum.

\section{EvoLLM with Teacher Fine-Tuning}



\begin{figure}
    \centering
    \includegraphics[width=0.325\textwidth]{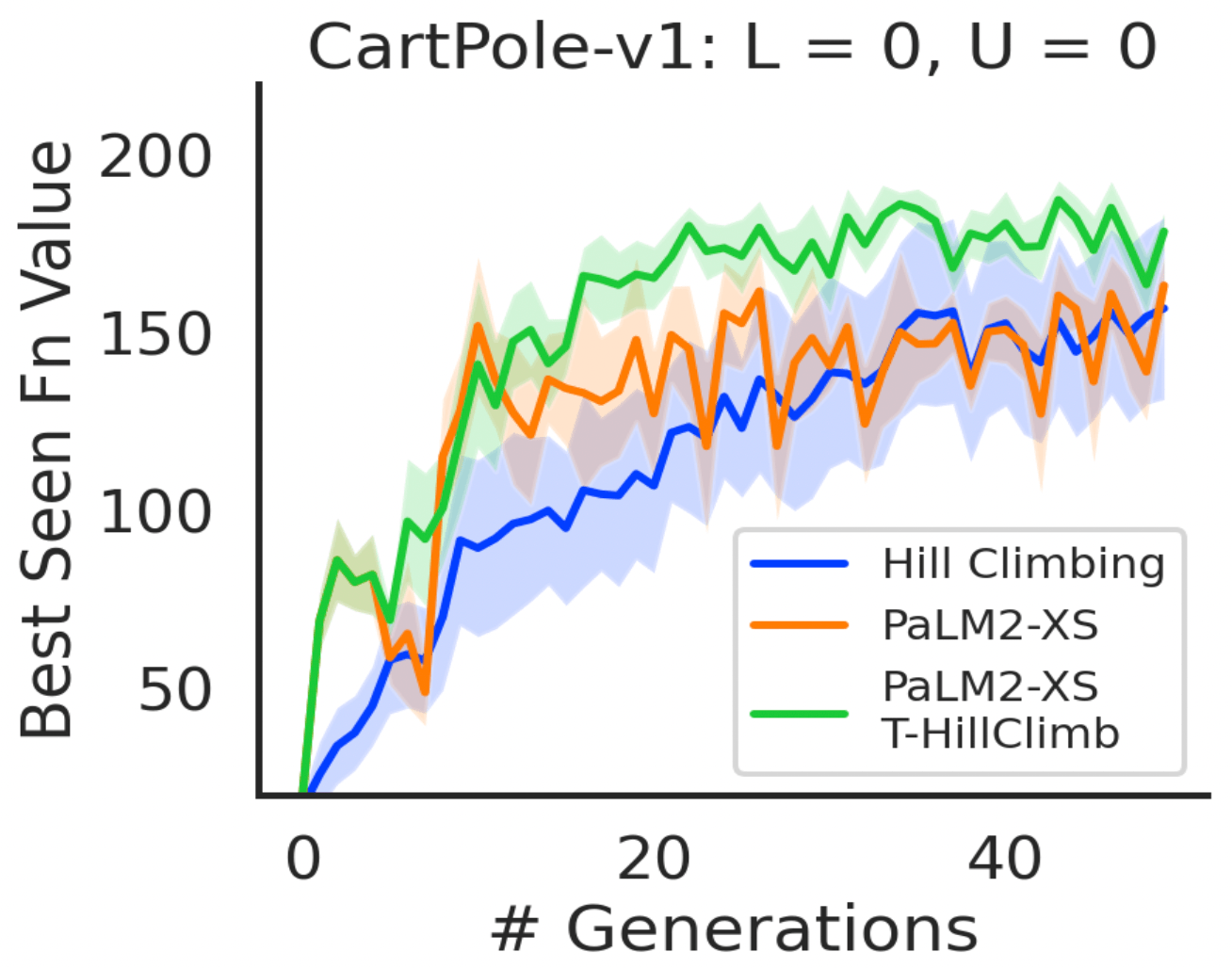}
    \caption{Instruction fine-tuning using Hill-Climbing improves EvoLLM's performance on unseen CartPole task.}
    \label{fig:results_finetune2}
\end{figure}

Finally, we investigate the impact of fine-tuning the LLM using BBO trajectories generated by a teacher algorithm. More specifically, we use two BBOB functions (Sphere and Rosenbrock) and collect BBO rollouts using a simple Gaussian Hill Climbing algorithm. Afterwards, we discretize the solution candidates and construct sequences of EvoLLM instructions according to the prompt template outlined in \cref{sec:evollm_prompt}.
We then train the LLM `backend' to predict the discretized integer representation of the teacher's search distribution update. To that end, we used the standard next-token prediction cross-entropy loss and fine-tuned it for a short amount of training steps (\cref{fig:supp_ftune}). Afterwards, we deploy the EvoLLM and evaluate its performance on the BBO tasks. Our results focus on the PaLM-XS model. 
\cref{fig:results_finetune} demonstrates the effect of fine-tuning on four different tasks (Sphere and Rastrigin with 2 and 5 search dimensions). For all cases, we observe a small but robust performance increase after instruction-based fine-tuning. In almost all considered cases, the tuned LLM outperforms both the teacher algorithm and the non-tuned LLM.
\cref{fig:results_finetune2} provides further evidence that this result extends also to more challenging neuroevolution tasks.
In summary, this highlights the potential for text-based LLMs to potentially distill teacher algorithms. One promising future direction may want to explore tuning the tokenizer vocabulary to better accommodate the representation of floating point numbers.

\section{Discussion}

\textbf{Summary}. We outline a prompt strategy that enables purely text-trained LLMs to robustly act as an ES on various BBO tasks. Furthermore, we provide several ablations highlighting the importance of careful solution representation and context construction. Finally, we successfully demonstrate that the LLMs capabilities can be improved by fine-tuning on teacher algorithm sequences.\\
\textbf{Limitations}. We expect EvoLLM's performance to differ based on pretraining \& fine-tuning protocols. Understanding how these details affect BBO performance is a key open challenge. Going forward LLMs capable of long context reasoning may be required to scale to larger search spaces. In preliminary experiments, we extended the prompt strategy to non-isotropic ES. The LLM had to additionally output an update to the diagonal covariance. This did not yield significant improvements. \\
\textbf{Future Work}. A potential direction is the construction of an LLM-specific BBO benchmark capturing various optimization abilities. This could enable directed progress for this specific LLM capability.
Furthermore, we want to explore tokenization techniques tailored for numerical representations. Finally, we expect EvoLLM to be able to harness all future advances of LLMs going forward, e.g. longer context windows.\\
\textbf{Ethical Considerations}. LLMs are powerful black-box tools that require careful monitoring of their agency. This is especially true when used for autonomous optimization purposes as the ones outlined in our work.

\newpage

\bibliographystyle{ACM-Reference-Format}
\renewcommand*{\bibfont}{\tiny}
\bibliography{bibliography}


\appendix

\newpage
\section{Additional Results}

\subsection{Impact of Search Space Resolution}
\begin{figure}[H]
    \centering
    \includegraphics[width=0.475\textwidth]{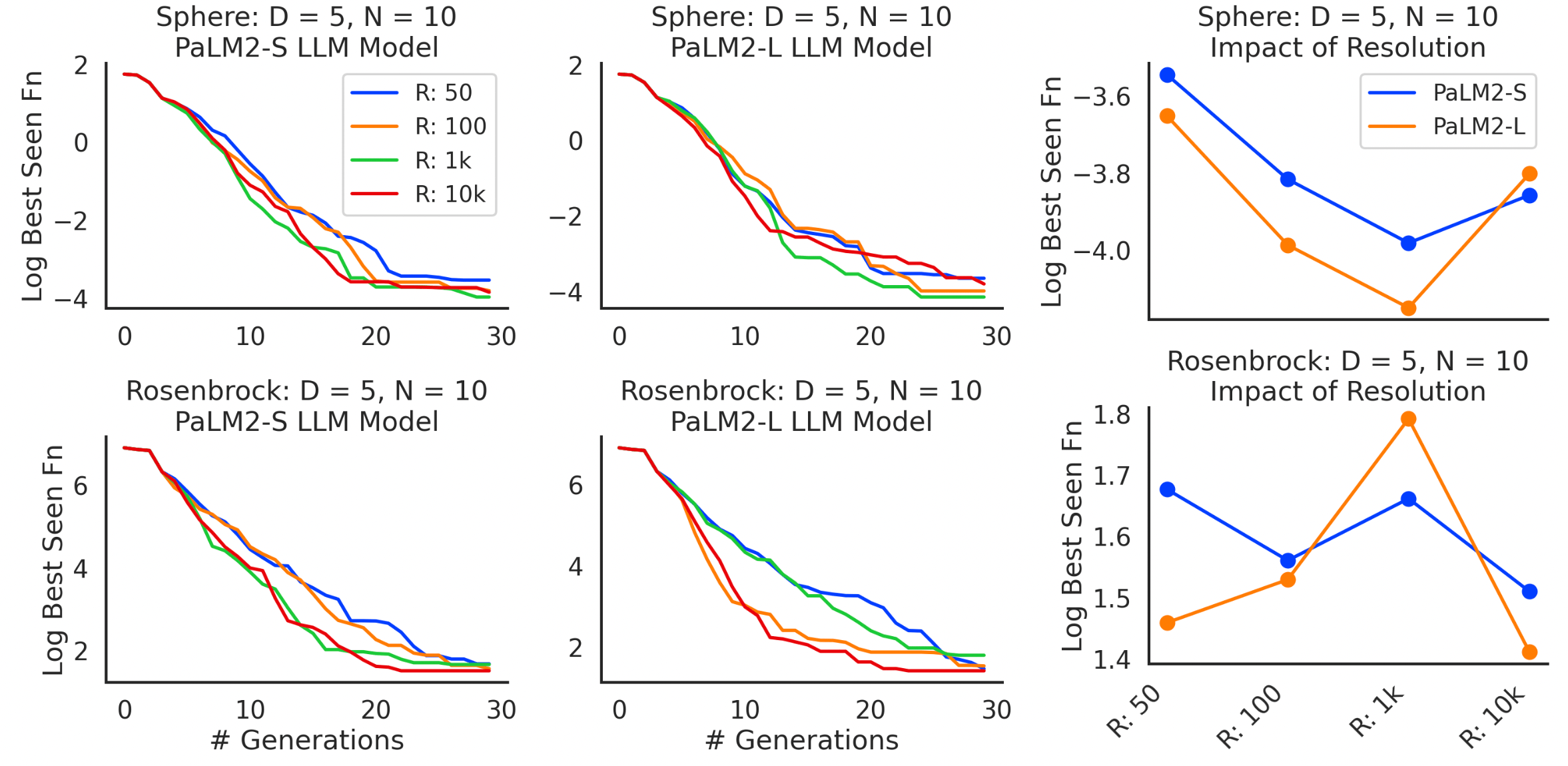}
    \caption{EvoLLM performance for different search resolutions. It struggles to zoom into optima basins for low-resolutions, while high resolutions degrade performance due to tokenization. Averaged results over 5 independent runs.}
    \label{fig:results_resolution}
\end{figure}

\subsection{Context Members \& Generations}
\begin{figure}[H]
    \centering
    \includegraphics[width=0.45\textwidth]{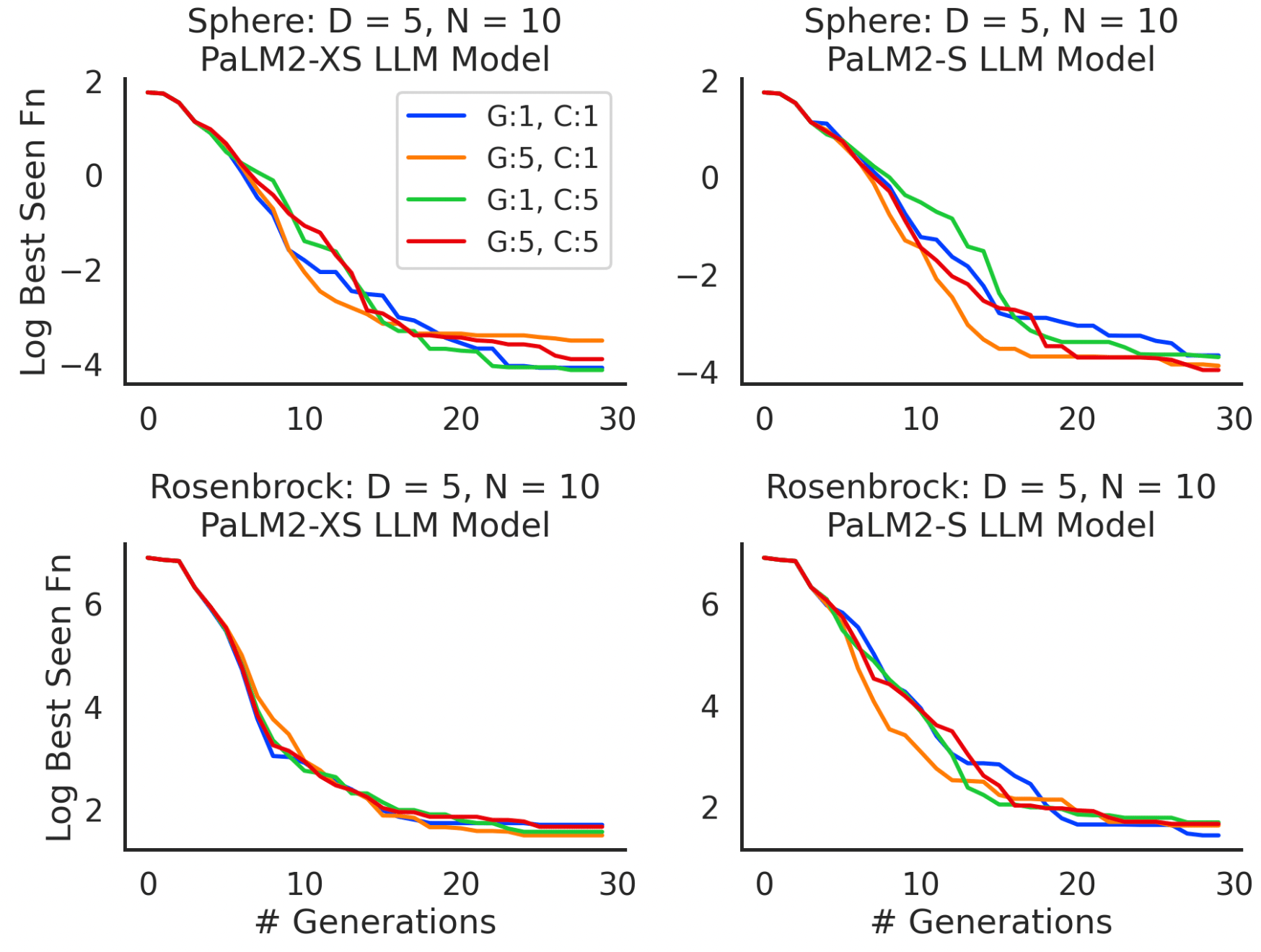}
    \caption{EvoLLM performance for different amounts of context information. Even with limited information the LLM is capable of inferring improving sequence patterns. Averaged results over 5 independent runs.}
    \label{fig:results_context_info}
\end{figure}

\section{Prompt Construction Example: 2 Dim. Sphere / 5 Population members}

Below we show an example of the EvoLLM prompt outlined in \cref{table:context_settings} for a 2-dimensional Sphere task with 5 population members. We start with a randomly initialized mean and 4 random search warm-up generations:

%
%
$\mu = [ 2.93851025 -1.77656265], f^\star = 0.4378929557544995$ 
\newpage
================ Generation 0 ================= \\
\begin{empheq}[box=\mymath]{align*}
\scriptstyle
&0.44: 397 539;147 92,0,0 302,0,186 346,0,419 685,0,397 539,0\\
&0.39: \Rightarrow \mu = [-0.61939515  0.2329004 ], \ f^\star = 0.438
\end{empheq}
================ Generation 1 =================\\
\begin{empheq}[box=\mymath]{align*}
&1.29: 397 539;559 140,0,186 346,0,419 685,0,670 417,0,397 539,0 \\
&0.44: 397 539;147 92,0,0 302,0,186 346,0,419 685,0,397 539,0 \\
&0.29: \Rightarrow \mu = [-0.61939515  0.2329004 ], \ f^\star = 0.438 
\end{empheq}
================ Generation 2 =================\\
\begin{empheq}[box=\mymath]{align*}
&6.03: 397 539;559 140,0,186 346,0,419 685,0,670 417,0,397 539,0 \\
&1.29: 397 539;559 140,0,186 346,0,419 685,0,670 417,0,397 539,0 \\
&0.44: 397 539;147 92,0,0 302,0,186 346,0,419 685,0,397 539,0 \\
&0.29: \Rightarrow  \mu = [-0.61939515  0.2329004 ], \ f^\star = 0.438
\end{empheq}
================ Generation 3 ================= \\
\begin{empheq}[box=\mymath]{align*}
&6.03: 397 539;559 140,0,186 346,0,419 685,0,670 417,0,397 539,0 \\
&2.48: 397 539;748 280,0,316 687,0,419 685,0,670 417,0,397 539,0 \\
&1.29: 397 539;559 140,0,186 346,0,419 685,0,670 417,0,397 539,0 \\
&0.44: 397 539;147 92,0,0 302,0,186 346,0,419 685,0,397 539,0 \\
&0.34: \Rightarrow  \mu = [-0.618  0.234], \ f^\star = 0.339
\end{empheq}
================ Generation 4 ================= \\
\begin{empheq}[box=\mymath]{align*}
&6.03: 397 539;559 140,0,186 346,0,419 685,0,670 417,0,397 539,0\\
&2.48: 397 539;748 280,0,316 687,0,419 685,0,670 417,0,397 539,0\\
&1.29: 397 539;559 140,0,186 346,0,419 685,0,670 417,0,397 539,0\\
&0.44: 397 539;147 92,0,0 302,0,186 346,0,419 685,0,397 539,0\\
&0.34: 413 543;388 557,0,448 604,0,397 539,0,399 504,1,413 543,1\\
&0.25: \Rightarrow \mu = [-0.522  0.258], \ f^\star = 0.249
\end{empheq}
================ Generation 5 ================= \\
\begin{empheq}[box=\mymath]{align*}
&6.03: 397 539;559 140,0,186 346,0,419 685,0,670 417,0,397 539,0 \\
&2.48: 397 539;748 280,0,316 687,0,419 685,0,670 417,0,397 539,0 \\
&1.29: 397 539;559 140,0,186 346,0,419 685,0,670 417,0,397 539,0 \\
&0.34: 413 543;388 557,0,448 604,0,397 539,0,399 504,1,413 543,1 \\
&0.25: 423 531;417 564,0,399 504,0,413 543,0,415 522,1,423 531,1 \\
&0.20: \Rightarrow \mu = [-0.462  0.186], \ f^\star = 0.15
\end{empheq}
================ Generation 6 ================= \\
\begin{empheq}[box=\mymath]{align*}
&6.03: 397 539;559 140,0,186 346,0,419 685,0,670 417,0,397 539,0 \\
&2.48: 397 539;748 280,0,316 687,0,419 685,0,670 417,0,397 539,0 \\
&0.34: 413 543;388 557,0,448 604,0,397 539,0,399 504,1,413 543,1\\
&0.25: 423 531;417 564,0,399 504,0,413 543,0,415 522,1,423 531,1\\
&0.15: 436 506;413 543,0,415 522,0,423 531,0,487 564,1,436 506,1\\
&0.09: \Rightarrow \mu = [-0.384  0.036], \ f^\star = 0.101
\end{empheq}
================ Generation 7 ================= \\
\begin{empheq}[box=\mymath]{align*}
&2.48: 397 539;748 280,0,316 687,0,419 685,0,670 417,0,397 539,0 \\
&0.34: 413 543;388 557,0,448 604,0,397 539,0,399 504,1,413 543,1 \\
&0.25: 423 531;417 564,0,399 504,0,413 543,0,415 522,1,423 531,1 \\
&0.15: 436 506;413 543,0,415 522,0,423 531,0,487 564,1,436 506,1 \\
&0.10: 447 499;434 527,0,487 564,0,436 506,0,439 514,1,447 499,1 \\
&0.07: \Rightarrow \mu = [-0.318 -0.006], \ f^\star = 0.040
\end{empheq}
================ Generation 8 ================= \\
\begin{empheq}[box=\mymath]{align*}
&0.34: 413 543;388 557,0,448 604,0,397 539,0,399 504,1,413 543,1 \\
&0.25: 423 531;417 564,0,399 504,0,413 543,0,415 522,1,423 531,1 \\
&0.15: 436 506;413 543,0,415 522,0,423 531,0,487 564,1,436 506,1 \\
&0.10: 447 499;434 527,0,487 564,0,436 506,0,439 514,1,447 499,1 \\
&0.04: 472 481;436 506,0,439 514,0,442 500,0,447 499,0,472 481,1 \\
&0.03: \Rightarrow \mu = [-0.168 -0.114], \ f^\star = 0.018
\end{empheq}
================ Generation 9 ================= \\
\begin{empheq}[box=\mymath]{align*}
&0.25: 423 531;417 564,0,399 504,0,413 543,0,415 522,1,423 531,1 \\
&0.15: 436 506;413 543,0,415 522,0,423 531,0,487 564,1,436 506,1 \\
&0.10: 447 499;434 527,0,487 564,0,436 506,0,439 514,1,447 499,1 \\
&0.04: 472 481;436 506,0,439 514,0,442 500,0,447 499,0,472 481,1 \\
&0.02: 478 495;465 474,0,472 481,0,476 519,1,479 485,1,478 495,1 \\
&0.01: \Rightarrow \mu = [-0.132 -0.03 ], \ f^\star = 0.018
\end{empheq}
================ Generation 10 ================= \\
\begin{empheq}[box=\mymath]{align*}
&0.15: 436 506;413 543,0,415 522,0,423 531,0,487 564,1,436 506,1 \\
&0.10: 447 499;434 527,0,487 564,0,436 506,0,439 514,1,447 499,1 \\
&0.05: 478 495;483 467,0,472 481,0,476 519,0,479 485,0,478 495,0 \\
&0.04: 472 481;436 506,0,439 514,0,442 500,0,447 499,0,472 481,1 \\
&0.02: 478 495;465 474,0,472 481,0,476 519,1,479 485,1,478 495,1 \\
&0.01: \Rightarrow \mu = [-0.132 -0.03 ], \ f^\star = 0.018
\end{empheq}

\newpage
\section{Floating Point Number Tokenization}

Below we show that the number of tokens used for a floating pointing number linearly increases with the number of digits. The numbers are sampled randomly:

\begin{figure}[H]
    \centering
    \includegraphics[width=0.35\textwidth]{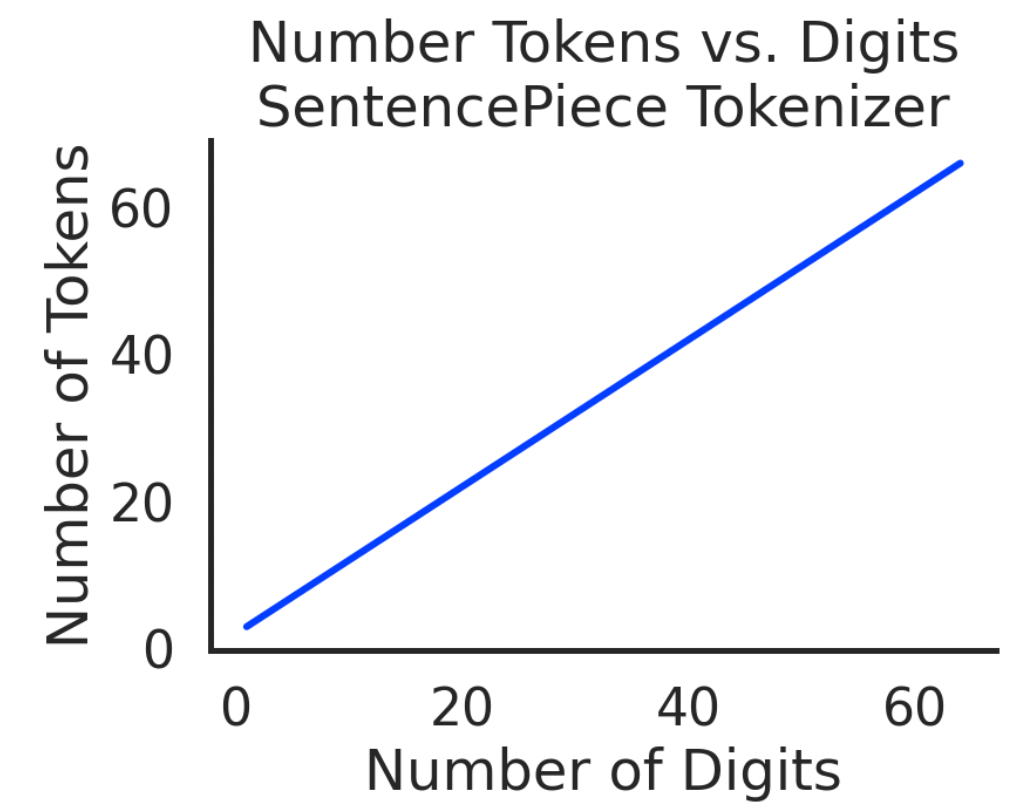}
    \caption{SentencePiece Tokenizer floating point number digits versus resulting tokens used for the language model.}
    \label{fig:supp_tokenizer}
\end{figure}

\section{Hill Climbing Instruction Fine-Tuning}

\begin{figure}[H]
    \centering
    \includegraphics[width=0.35\textwidth]{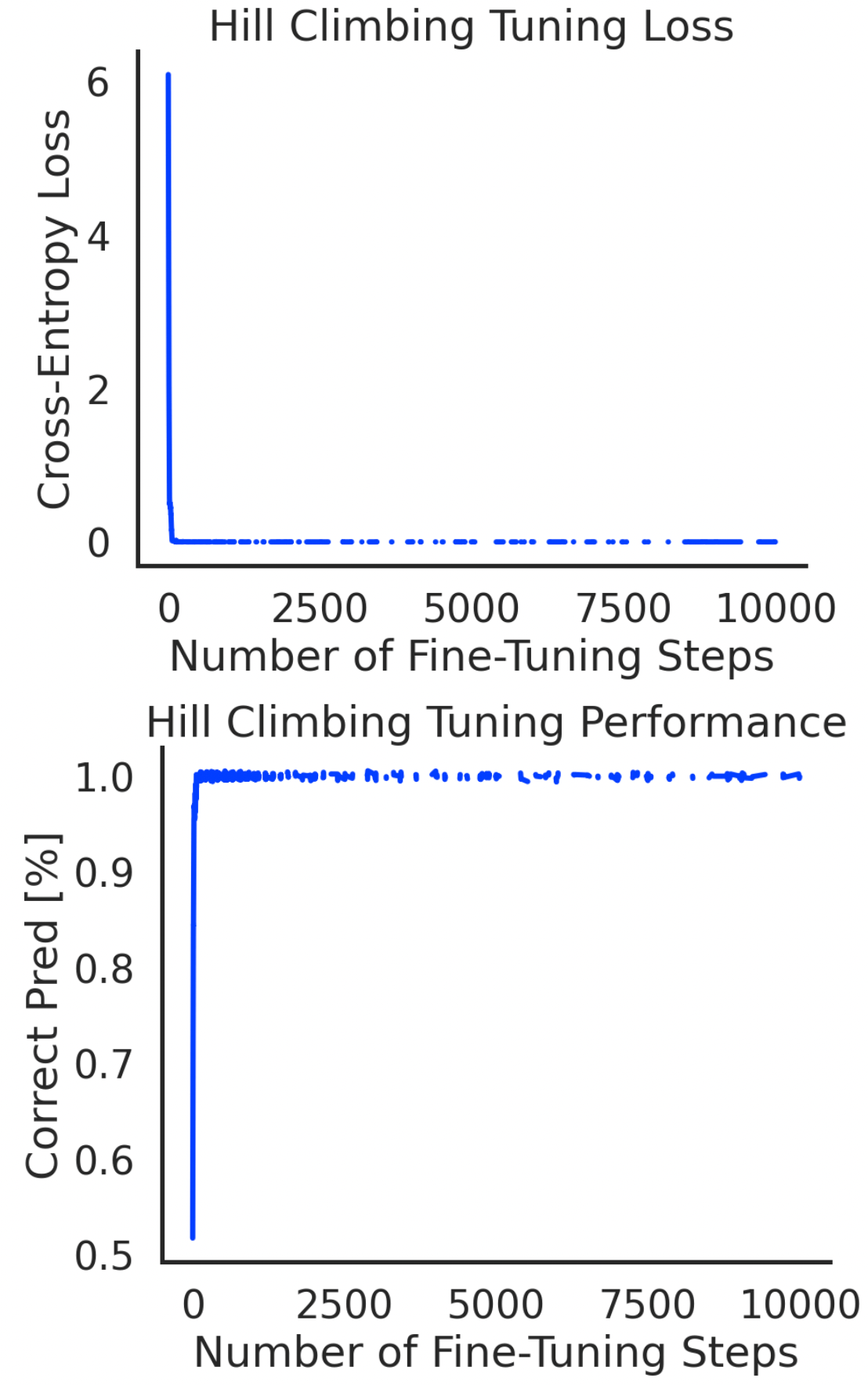}
    \caption{Fine-tuning loss and accuracy for instruction fine-tuning using Hill Climbing teacher algorithm trajectories on PaLM-XS.}
    \label{fig:supp_ftune}
\end{figure}

\end{document}